\documentclass[journal=acscii,manuscript=article]{achemso}

\usepackage[version=3]{mhchem} 

\usepackage{xcolor}
\usepackage{color}

\usepackage[font=footnotesize,labelfont=bf]{caption}

\usepackage{lscape}
\usepackage{fullpage}
\usepackage{multicol,multirow}
\usepackage{booktabs}
\usepackage{graphicx}

\usepackage{algpseudocode,algorithm}
\usepackage{algorithm}
\usepackage{algpseudocode}
\usepackage{verbatim}
\usepackage{ulem}
\usepackage{arydshln}

\author{Zhongliang Guo}
\affiliation[ISM]
{The Institute of Statistical Mathematics, Research Organization of Information and Systems, Tachikawa, Tokyo 190-8562, Japan}

\author{Stephen Wu}
\affiliation[ISM]
{The Institute of Statistical Mathematics, Research Organization of Information and Systems, Tachikawa, Tokyo 190-8562, Japan}
\alsoaffiliation[SOKENDAI]
{The Graduate University for Advanced Studies, SOKENDAI, Tachikawa, Tokyo 190-8562, Japan}

\author{Mitsuru Ohno}
\affiliation[DAICEL]
{Daicel Corporation, Kita-ku, Osaka 530-0011, Japan}

\author{Ryo Yoshida}
\affiliation[ISM]
{The Institute of Statistical Mathematics, Research Organization of Information and Systems, Tachikawa, Tokyo 190-8562, Japan}
\alsoaffiliation[SOKENDAI]
{The Graduate University for Advanced Studies, SOKENDAI, Tachikawa, Tokyo 190-8562, Japan}
\alsoaffiliation[NIMS]
{National Institute for Materials Science, Tsukuba, Ibaraki 305-0047, Japan}
\email{yoshidar@ism.ac.jp}

\title[Bayesian retrosynthesis]
  {A Bayesian algorithm for retrosynthesis} 

\abbreviations{}

\begin{document}

\makeatletter
\setlength\acs@tocentry@height{8.75cm}
\setlength\acs@tocentry@width{2.5cm}
\makeatother

\begin{tocentry}
\begin{center}
    \includegraphics{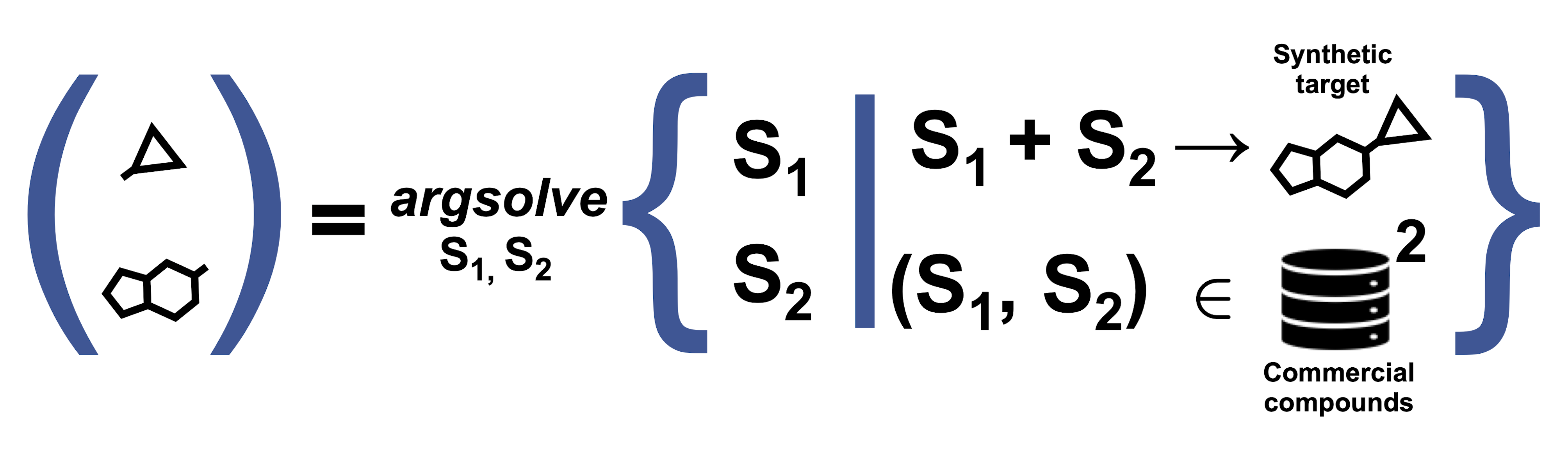}
\end{center}
\end{tocentry}

\begin{abstract}
The identification of synthetic routes that end with a desired product has been an inherently time-consuming process that is largely dependent on expert knowledge regarding a limited fraction of the entire reaction space. At present, emerging machine-learning technologies are overturning the process of retrosynthetic planning. The objective of this study is to discover synthetic routes backwardly from a given desired molecule to commercially available compounds. The problem is reduced to a combinatorial optimization task with the solution space subject to the combinatorial complexity of all possible pairs of purchasable reactants. We address this issue within the framework of Bayesian inference and computation. The workflow consists of two steps: a deep neural network is trained that forwardly predicts a product of the given reactants with a high level of accuracy, following which this forward model is inverted into the backward one via Bayes' law of conditional probability. Using the backward model, a diverse set of highly probable reaction sequences ending with a given synthetic target is exhaustively explored using a Monte Carlo search algorithm. The Bayesian retrosynthesis algorithm could successfully rediscover 80.3\% and 50.0\% of known synthetic routes of single-step and two-step reactions within top-10 accuracy, respectively, thereby outperforming state-of-the-art algorithms in terms of the overall accuracy. Remarkably, the Monte Carlo method, which was specifically designed for the presence of diverse multiple routes, often revealed a ranked list of hundreds of reaction routes to the same synthetic target. We investigated the potential applicability of such diverse candidates based on expert knowledge from synthetic organic chemistry.
\end{abstract}

\section{Introduction}
The objective of retrosynthetic planning is to design a synthetic route backwardly from a given desired molecule to commercially available starting materials. In 1969, Corey and Wipke introduced the first computer-aided synthesis planning program, known as Organic Chemical Simulation of Synthesis \cite{Corey1969Computer-AssistedSyntheses}, which has continued to grow as LHASA \cite{PENSAK1977}, SYNCHEM \cite{Gelernter1977EmpiricalSYNCHEM} and WODCA \cite{Gasteiger2000Computer-assistedChemistry}, among others. Such early retrosynthetic systems relied on hand-coded reaction rules or those algorithmically extracted from reaction databases. The applicability of a feasible reaction rule to a target product is assessed based on the presence of local structural or atomic features around the candidate reaction site in the rule set. Knowledge-based or, more recently, machine-learning algorithms have been used to judge which rule to select. Such prioritized transformation, for example, a rule to break bonds, is recursively applied to the current molecules to derive structurally simpler precursors until the growing retrosynthetic tree reaches readily available substrates.

Rule-based systems are interpolative by nature. Their predictability is no longer applicable if the underlying reaction mechanisms extend beyond the synthetic knowledge encoded in the rule set. To cover a broader reaction space, various deep-learning techniques have been introduced to retrosynthetic analyses in recent years, which can be categorized into rule-based two-step models \cite{Segler2017Neural,Coley2017Computer-AssistedSimilarity} and fully data-driven end-to-end analyses \cite{Liu2017RetrosyntheticModels,Zheng2019PredictingNetworks,Lin2019AutomaticModels}. Their common goal is to identify the inverse mapping of a synthetic reaction from any given product to its unknown reactants using millions of training examples from known reactions. The former methods divide the task into two separate steps: the use of a trained machine-learning model that classifies an input product into one of dozens or hundreds of pre-compiled reaction templates, followed by the application of rules that are more likely to occur to deconvolve the target product retrosynthetically into simpler precursors \cite{Segler2017Neural,Coley2017Computer-AssistedSimilarity}. The ability to predict the reaction outcomes has been significantly improved compared to earlier knowledge-based or logic-based models, owing to the high coverage of training reaction instances. However, these models still rely on reaction rules, resulting in limitations in the coverage of the predictable reaction spaces. To broaden the search space further, fully end-to-end approaches based on state-of-the-art neural machine translation systems have been developed, such as seq2seq \cite{Liu2017RetrosyntheticModels} and Molecular Transformer \cite{Schwaller2019MolecularPrediction}. Once the chemical structures of products and reactants have been encoded by the Simplified Molecular Input Line Entry Specification (SMILES) chemical language \cite{Weininger1988SMILESRules}, the task of predicting the retrosynthetic outcomes amounts to determining the rule of conversion from character-encoded products to character-encoded reactants.
With a given synthetic target, such backward reaction prediction models can be used to generate a branched sequence of reactants recursively, until the growing retrosynthetic tree reaches a prescribed set of purchasable compounds. Several general-purpose best-first search algorithms can be used to identify chemically plausible synthetic routes from a potentially infinitely
large search tree effectively, such as Monte Carlo tree search \cite{Segler2018PlanningAI}. 

It should be noted that the majority of candidate reactants simulated from such backward prediction models are rarely contained within a given set of purchasable compounds that span the feasible solution space. This is a major drawback of existing retrosynthetic methods. Another shortcoming arises from the low accuracy of backward reaction models. One reason for this is the loss of information in reaction data; the side products of synthetic reactions are often unrecorded in the datasets. To predict the reactants in a backward manner, it is necessary to reconstruct the missing structure of their product, which is impossible to achieve without using additional knowledge on structural transformation. As summarized in Table \ref{table:table1}, the previously reported prediction accuracy ranged from 35\% to 45\%. This implies that a large number of candidate reactants in a simulated reaction sequence would be commercially unobtainable.

However, the forward prediction of reaction outcomes has achieved substantially higher levels of accuracy than those of retrosynthetic prediction \cite{Coley2017PredictionLearning,Jin2017PredictingNetwork,Schwaller2019MolecularPrediction}. As summarized in Table \ref{table:table1}, the reported prediction accuracy of state-of-the-art models is 90.4\% for Molecular Transformer, which is twice the accuracy of backward prediction.
The key concept of this study is rather obvious. A trained forward model with such high accuracy defines the mapping $Y = f(S)$ from a set of reactants $S$ to their product $Y$. By solving the inverse mapping $S = f^{-1}(Y^*)$ with a synthetic target $Y^*$ with respect to a pre-defined pool $\mathcal{S}$ of commercially available reactants, there is a chance that the resulting retrosynthetic prediction will reach the same level of accuracy as that of the forward prediction model. The problem to be solved is a combinatorial optimization problem with the solution space subject to the combinatorial complexity of all possible pairs of purchasable reactants in the catalog. The complexity increases exponentially with the size of the candidate reactants, as well as the number of reaction steps considered. 

In this study, we addressed the task of reaction mining within the framework of Bayesian inference, namely Bayesian retrosynthesis, which provides a principled means of inverting any given forward models into the retrosynthetic prediction system. To enhance the search efficiency and exhaustively enumerate alternative pathways, we developed a sequential Monte Carlo (SMC) algorithm using a surrogate model accelerator (see Figure 1 for a schematic view). 

The Bayesian retrosynthesis algorithm successfully rediscovered 80.3\% and 50.0\% of known synthetic routes of single-step and two-step reactions within top-10 accuracy, respectively.
Remarkably, the Monte Carlo method, which was specifically designed for the presence of diverse multiple routes, revealed a prioritized list of more than 400 reaction routes on average for each target. We investigated the potential applicability of such diverse candidates based on expert knowledge from synthetic organic chemistry. A Python implementation is available on GitHub \cite{Guo2020BayesianRetro}.

\begin{figure}
  \centering
  \includegraphics[clip, width=16.5cm]{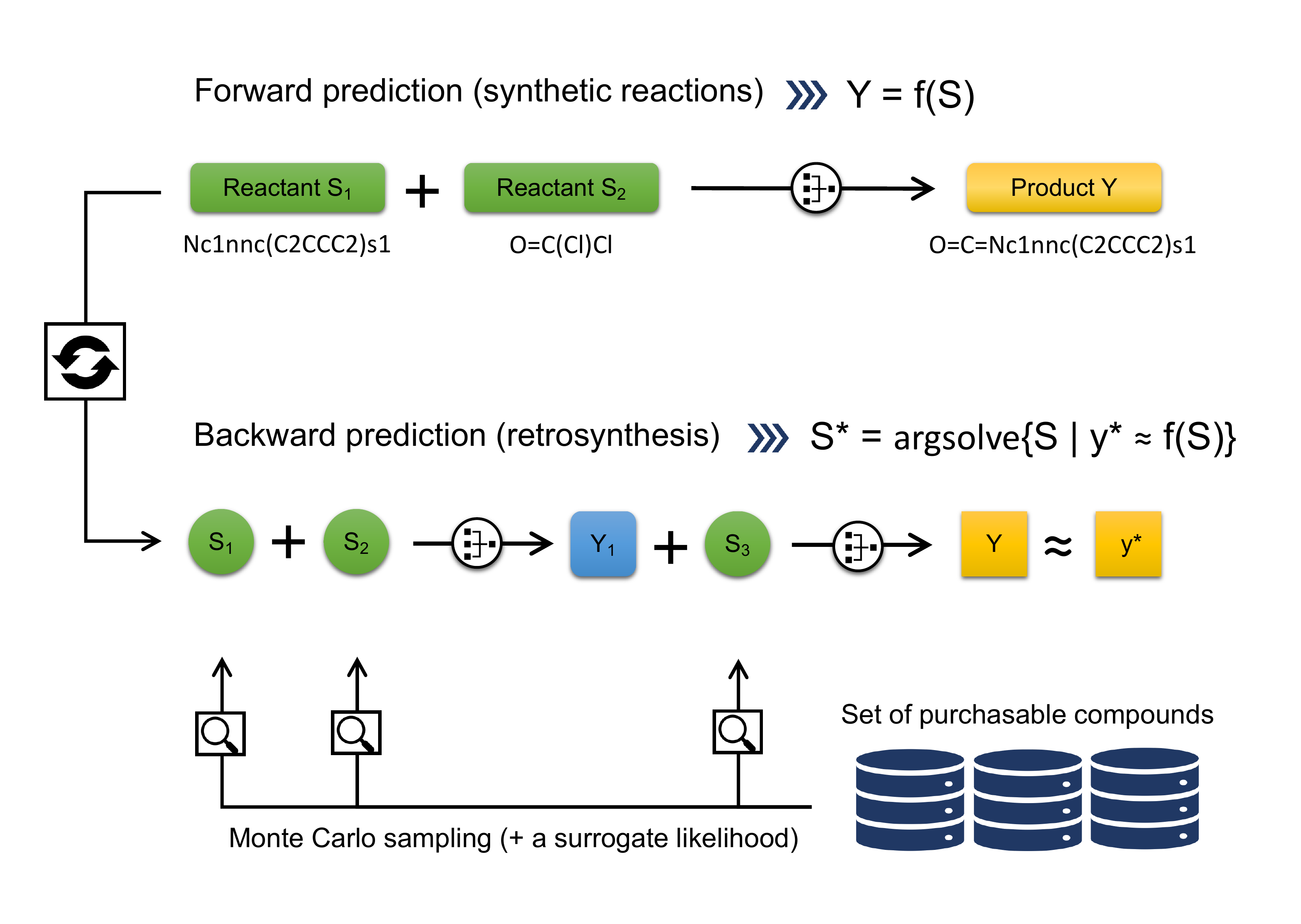}
  \caption{\textcolor{black}{Workflow of Bayesian retrosynthesis algorithm. A synthetic reaction model $Y=f(S)$ that forwardly predicts a product $Y$ of any given reactants $S$ is inverted into the backward model through Bayes' law of conditional probability. Monte Carlo sampling from the posterior distribution conditioned by a desired product $Y=y^*$ provides a diverse set of highly probable $M$ reactant pairs, $\mathcal{P} = \{S_1, \ldots, S_M\}$, which meet the requirement $y^* \simeq f(S)$.}}
  \label{fgr:fig1}
\end{figure}

\begin{table}[t]
\centering
\small
\begingroup
\renewcommand{\arraystretch}{1.3}
\textsf{
\begin{tabular}{l l c c c c} \toprule
Task & Model & top-1 & top-3 & top-5 & top-10 \\ \hline
\multirow{3}{*}{Backward} & similarity (Coley et al. 2017) \cite{Coley2017Computer-AssistedSimilarity} & 37.3 & 54.7 & 63.3 & 74.1 \\
 & SCROP (Zheng et al. 2019) \cite{Zheng2019PredictingNetworks} & 43.7 & 60.0 & 65.2 & 68.7 \\
 & Lin et al. 2019 \cite{Lin2019AutomaticModels} & 43.1 & 64.6 & 71.8 & 78.7 \\ \hline
 \multirow{3}{*}{Forward} & template-based (Coley et al. 2017) \cite{Coley2017PredictionLearning} & 71.8 & 86.7 & 90.8 & 94.6 \\
  & WLDN (Jin et al. 2017) \cite{Jin2017PredictingNetwork} & 79.6 & 87.7 & 89.2 & - \\
  & Molecular Transformer (Schwaller et al. 2019) \cite{Schwaller2019MolecularPrediction} & 90.4 & 94.6 & 95.3 & - \\
\bottomrule
\end{tabular}
}
\endgroup
    \caption{\footnotesize \textcolor{black}{Performance of existing deep neural networks on prediction of synthetic reactions in forward and backward manners. The top-1, top-3, top-5, and top-10 accuracies in [\%] are presented for each.}}
    \label{table:table1}
\end{table}

\section{Methods}

\subsection{Outline}

A single-step reaction prediction model $f(S)$ is a function that describes a product $Y$ as a function of a set of reactants, denoted by $S$. Using such a model, we can simulate a single-step reaction for any $S$ as $S \rightarrow Y$. Herein, the function $f$ is treated as being deterministic. Solvents, reagents, and catalysts can be augmented into the input variable as required.

Likewise, a $k$-step reaction sequence ending with a final product $Y$ can be generated by convoluting the single-step model $f(S)$ $k$ times with arbitrarily selected reactant sets $S_1, \ldots, S_k$ at each step:
\begin{eqnarray}
S_1 \rightarrow 
Y_1 + S_2 \rightarrow
\ldots
\rightarrow
Y_{k-1} + S_{k} \rightarrow
Y, \nonumber
\end{eqnarray}
where $Y_i = f(Y_{i-1}, S_i)$ denotes an intermediate product that is produced from the product $Y_{i-1}$ at the previous step and the currently selected reactants $S_i$. In general, the reaction prediction can be represented as $Y = f(S)$, with all reactants concatenated into a single input symbol as $S = S_1 + \ldots + S_k$.

The ultimate goal of the retrosynthetic prediction is to enumerate all possible $S$ satisfying $y^* = f(S)$ for a given synthetic target $Y = y^*$, or equivalently, to solve the inverse mapping of the forward model $S = f^{-1}(y^*)$. The solution space $\mathcal{S}$ is composed of all reactant combinations in the purchasable compounds. The number of candidate reactants is typically of the order $O(10^6)$, resulting in the cardinality of the solution space being exploded to $|\mathcal{S}| = O(10^{6 \times r})$, in which $r$ reactants are placed in the synthetic route planning.

In certain cases, we aim to identify an ensemble of $S$ that meets the requirement approximately as $y^* \simeq f(S)$, rather than obtaining the strict solution. Firstly, it is possible that all candidate reactants will never reach the target product with the currently provided model $f(S)$. Furthermore, even when the model is incorrect, true reactants are expected to be near the optimal solutions $y^* \simeq f(S)$. In such scenarios, it is more beneficial to view the distribution of $S$ as approximately satisfying $y^* \simeq f(S)$ than to obtain only the strict solution. This is the underlying concept for handling the retrosynthetic analysis within the Bayesian framework.

The Bayesian retrosynthesis relies on Bayes' law of conditional probability:
\begin{eqnarray}
p(S | Y = y^*) \propto p(Y = y^*| S) p(S). \nonumber
\end{eqnarray}
This law states that the posterior probability distribution $p(S|Y = y^*)$ is proportional to the product of the likelihood $p(Y = y^*|S)$ and prior $p(S)$. The forward prediction model forms the likelihood function, which is the Boltzmann distribution with an inverse temperature $\beta$, as follows:
\begin{eqnarray}
p(Y = y^*| S) \propto \exp\Big\{ -\beta E(y^*, f(S)) \Big\}. \nonumber
\end{eqnarray}
The energy function $E$ is provided by the Euclidean distance or Tanimoto distance \cite{tanimoto1958elementary} between the chemical structures of the target $Y = y^*$ and predicted product $f(S)$. The distance is calculated with the extended-connectivity fingerprint (ECFP)\cite{Rogers2010Extended-connectivityFingerprints} with diameter 4 using the open-source cheminformatics toolkit RDKit \cite{landrum2006rdkit} in Python. Through the prior distribution, it is possible to place a low probability mass on unlikely reactants to narrow down the enormous solution space. We assigned equal probabilities to all candidates throughout this study so that $p(S) = \mathrm{const.}$ for all $S$.

The posterior is a discrete probability distribution $p(S|Y = y^*) \propto \sum_{i=1}^{|\mathcal{S}|} p(Y = y^*| S) \delta_{s_i}(S)$, where $\delta_{s_i}(S)$ denotes the indicator function that takes the value 1 if $S = s_i$, and 0 otherwise. The support of this discrete measure consists of all possible combinations of $r$ reactants involved in a synthetic route. As the exact computation across $O(10^{|\mathcal{S}| \times r})$ candidates is generally infeasible, we explore the approximated form $\hat{p}(S|Y = y^*)$ as
\begin{eqnarray}
\hat{p}(S|Y = y^*) \propto \sum_{i=1}^n p(Y = y^*| S) \delta_{s_i}(S). \label{eq:1}
\end{eqnarray}
The primary objective in the Bayesian computation is to identify the reduced set of $n$ reactant pairs, $\{s_i\}_{i=1}^n$, possibly with $n \ll |\mathcal{S}|$. Candidates with greater likelihoods $p(Y = y^*| s_i)$ should have a better chance of surviving, while ignorable $s_i$ with low likelihood values are effectively eliminated.

\subsection{Difficulties of ordinal heuristic algorithms}

Sampling from the posterior distribution over the huge discrete space is a considerably difficult task. Indeed, a conventional heuristic algorithm is no longer applicable to solve this task, as demonstrated in this study (see Supporting Information and Table S6). To observe this, we first present the simplest form of the sequential Monte Carlo (SMC) algorithm \cite{Moral2006SequentialSamplers}. The SMC has a common algorithmic structure with genetic algorithms, which has been applied in various machine-learning or signal-processing problems. It begins with an arbitrary set of $p$ particles, $\mathcal{P}_0 = \{s_i^0\}_{i=1}^{p}$, which represents an ensemble of candidate reactant pairs, followed by the iteration of two operations, hereafter referred to as {\it sampling} and {\it resampling}.

\begin{algorithm}
\caption{Simple SMC} \label{simpleSMC}
\begin{algorithmic}[1]
\State Initialize a set of $p$ particles $\mathcal{P}_0 = \{s_i^0\}_{i=1}^{p}$.
\For{$t = 1, \ldots, T$}
\For{$i = 1, \ldots, p$}
\State Generate a new particle $s_i^{*} \sim g(s_i^{*}| s_i^{t-1})$ according to a proposal distribution $g$.
\State Evaluate the importance weight according to the likelihood function
\begin{eqnarray}
w_i = \frac{p(Y = y^* | S = s_i^{*})}{ g(s_i^{*}| s_i^{t-1}) }. \nonumber
\end{eqnarray}
\EndFor
\State Resample $\{s_i^{*}\}_{i=1}^{p}$ with the probability proportional to $\{w_i\}_{i=1}^p$ to obtain a new set $\mathcal{P}_t = \{s_i^{t}\}_{i=1}^{p}$.
\EndFor
\end{algorithmic}
\end{algorithm}

To reveal the technical difficulties, we consider a simple SMC, as indicated in Algorithm \ref{simpleSMC}. For a given particle set $\mathcal{P}_{t-1}$ at step $t-1$, a particle $s_i^{t-1}$ is tentatively replaced by a new $s_i^{*}$ according to a prescribed proposal distribution $g(s_i^{*}| s_i^{t-1})$. Specifically, we sample $s_i^{*}$ with equal probability from the $k$-nearest neighbors of $s_i^{t-1}$ in the candidate set $\mathcal{S}$ of commercial reactants. For the neighbor search, all compounds in $\mathcal{S}$ and reactants in $s_i^{*}$ are fingerprinted by ECFP with diameter 4, followed by the calculation of the Tanimoto distance. For each candidate particle, the goodness-of-fit $w_i$ to the synthetic target, which is referred to as the importance weight, is evaluated by simulating the product using Molecular Transformer. It should be noted that the importance weight is reduced to $w_i \propto p(Y = y^* | S = s_i^{*})$ when using the $k$-nearest neighbor proposal $g(s_i^{*}| s_i^{t-1}) = 1/k$. Resampling of $\{s_i^{*}\}_{i=1}^p$ is then carried out based on the selection probability proportional to $\{w_i\}_{i=1}^p$, which yields a new particle set $\mathcal{P}_{t} = \{s_i^{t}\}_{i=1}^p$. For a given $s_i^t$, as the predicted product moves closer to the given synthetic target, its reactant set has a greater chance of surviving in the resampling step. By repeating the operation of sampling and resampling $T$ times, we obtain $p \times T$ samples in $\mathcal{P}_1, \ldots, \mathcal{P}_T$ with the likelihood $p(Y=y^*|S = s_i)$, which form the approximated posterior (Eq. \ref{eq:1}). 

As described in Supporting Information, this simple SMC performed exceptionally poorly, as it failed to discover approximately 50\% of known reactions in the performance test (Table S6), owing to two essential drawbacks that are common to ordinal heuristic search methods, including the genetic algorithm. One difficulty arose from the underlying diversity of the solutions. Our analysis demonstrated that a large number of synthetic routes ended with the same product in the search space.
Using an extended algorithm, more than 400 different routes were identified for the synthetic targets on average.
In general, it is difficult to identify a diverse set of highly probable solutions comprehensively using ordinal heuristic methods. In the SMC workflow, the step of repeated resampling induces a rapid loss of diversity among the particle sets, which is known as the problem of particle impoverishment \cite{Stavropoulos2001ImprovedSmoothing}. Another issue of the ordinal methods arose from the cost of the repeated calculations of reaction prediction models. On average, the single-step reaction prediction using Molecular Transformer required 30 to 40 seconds on a Linux server with a NVIDIA V100 GPU or P100 GPU, thereby leading to a total execution time of over 7 hours under $p = 1,000$ and $T = 600$. In summary, a surrogate model-assisted Monte Carlo algorithm was required to save the costs of repeatedly evaluating the computationally expensive reaction prediction model while maintaining the diversity of the particles, including highly probable solutions.

\subsection{SMC accelerated by surrogate likelihood}
\begin{figure}
  \centering
  \includegraphics[clip, width=16.5cm]{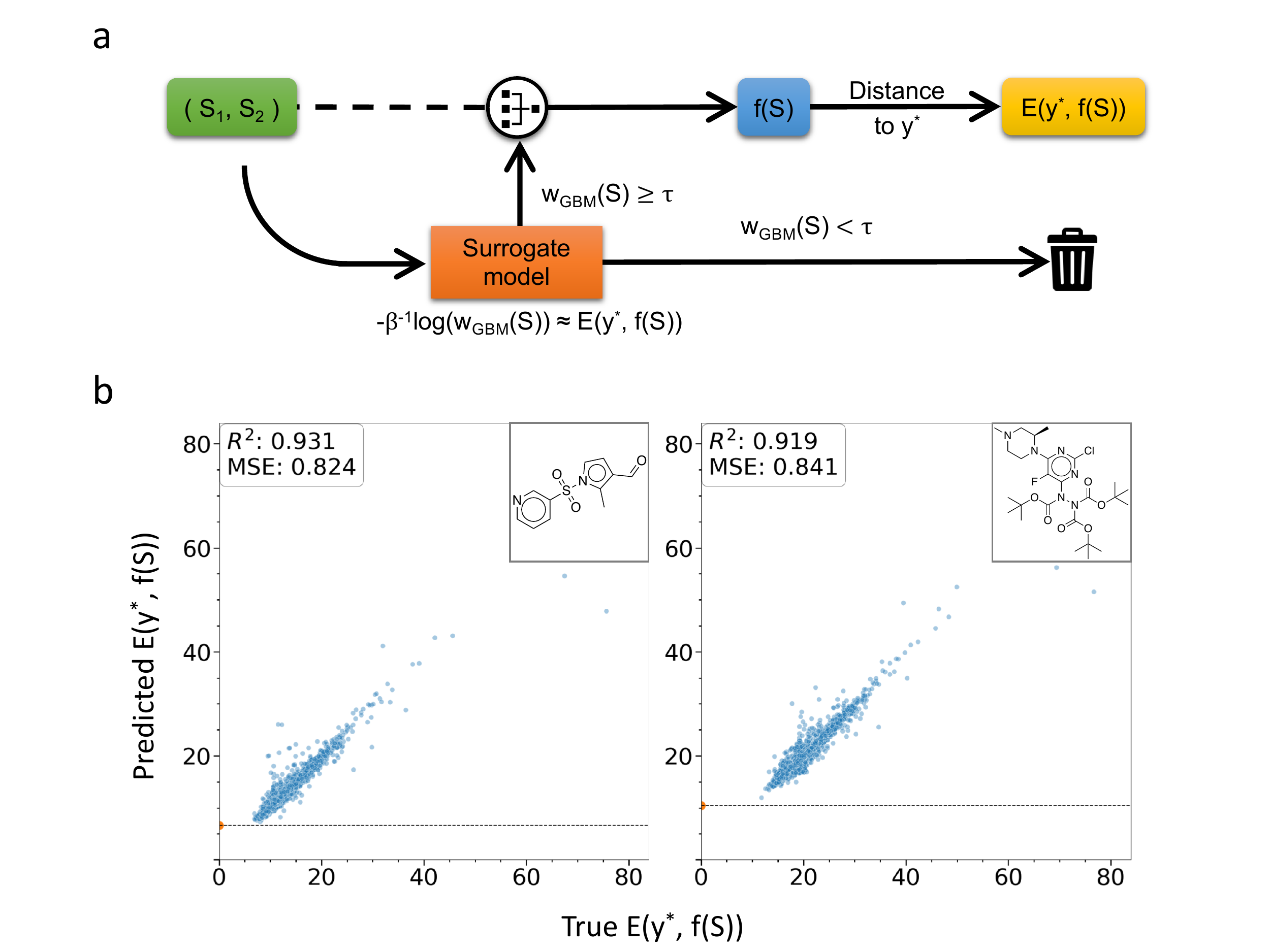}
  \caption{\textcolor{black}{{\textbf a.} The energies (negative log-likelihood) predicted by the surrogate model are used to prioritize promising reactants before conducting expensive reaction prediction using Molecular Transformer. {\textbf b.} For two synthetic targets, the energies predicted by the surrogate model are displayed against the true values of 5k reactions in the test dataset. The orange points denote the ground-truth reactants. The dashed lines indicate the predicted values of the ground-truth reactants. The target molecules are displayed in the top right corners.}}
  \label{fig:fig2}
\end{figure}

A key concept behind our strategy is that, by using a computationally inexpensive surrogate model, such as gradient boosting regression \cite{Schapire2012BoostingAlgorithms}, we can approximately evaluate the likelihood function of Molecular Transformer for any given reactants with a high degree of accuracy. For $m$ instances $\{s_i\}_{i= 1}^m$ of reactants, the likelihood values $w_i = p(Y=y^* | S = s_i)$ ($i = 1, \ldots, m$) are calculated using Molecular Transformer with any given target $y^*$. Using this dataset, we train a gradient boosting regression tree $w_{\mathrm{GBM}}(S)$, which can be used to predict the likelihood of $S$ without passing through Molecular Transformer (Figure \ref{fig:fig2}a). In this study, a gradient boosting model was pre-trained on the training dataset.
The ECFPs in the RDKit package were used as inputs for the gradient boosting regression based on the LightGBM package \cite{Ke2017LightGBM:Tree} in Python. Figure \ref{fig:fig2}b presents the performance of surrogate models for two different target molecules. The true reactants were scored as the most likely for each target molecule among the test data. The high correlation of the predicted and true likelihood made the surrogate likelihood a reliable alternative for the true likelihood, the calculation of which was expensive.

The proposed retrosynthesis algorithm is described in Algorithm \ref{surrogateSMC}. It relies on the evolutionary strategy involving the sampling and resampling operations, as in Algorithm \ref{simpleSMC}. Three modifications are included in the sampling step to maintain the diversity of the final candidates. Firstly, the currently provided top-$l$ particles are expanded into $l \times m$ particles by taking the $m$-nearest neighbors of each candidate (line 7). This operation aims to increase the heterogeneity of the sample population in the following generation. The fast-to-evaluate surrogate model can efficiently prioritize increasingly heterogeneous particles according to the surrogate likelihoods (line 8). Moreover, the $l \times m$ particles are clustered into $K$ subgroups according to the fingerprints (ECFP with diameter 4) of the chemical structures (line 9). The surrogate likelihoods are transformed into the cluster-level goodness-of-fit scores by taking the within-cluster average (line 10). As a result, a new set of particles is created by performing resampling of the $l \times m$ particles, such that the number of particles in each cluster is proportional to the within-cluster likelihood (line 11). This cluster-level resampling is key to maintaining the population diversity. Finally, a set of randomly selected candidate reactants is augmented to the new set of particles (line 12). After calculating the true likelihoods of the surviving particles using Molecular Transformer, we proceed with the selection of the top-$l$ particles.

\begin{algorithm}
\caption{Surrogate-assisted SMC} \label{surrogateSMC}
\begin{algorithmic}[1]
\State Initialize a set of $p$ particles: $\mathcal{P}_0 = \{s_i^0\}_{i=1}^{p}$ with likelihood $\mathcal{W}_0 = \{w_{i}^0\}_{i=1}^p$.
\State Initialize the active candidate set $\mathcal{A} = \{1, \ldots, n\}$, which consists of the indices of all possible reactant pairs.
\State Generate a training set $\mathcal{D} = \{s_i, w_i\}_{i = 1}^n$ to construct a surrogate likelihood function, in which Molecular Transformer is used to provide the likelihood $w_i$ for any $s_i$.
\State Build a pre-trained model (gradient boosting) on $\mathcal{D}$ as the surrogate likelihood $w_{\mathrm{GBM}}(S)$.
\For{$t = 1, \ldots, T$}
\State Select the top $l$ of $p$ particles from $\mathcal{P}_{t-1}$ according to the descending order of likelihood values in $\mathcal{W}_t$.
\State Generate $l \times m$ particles $\mathcal{P}^{*} = \{s_i^{*}\}_{i=1}^{l \times m}$ by taking the $m$-nearest neighbors of each $s_i^{t-1}$ ($i=1, \ldots, l$) selected from $\mathcal{A}$.
\State Calculate the surrogate likelihood $w_{\mathrm{GBM}}(s_i^{*})$ of $s_i^{*} \in \mathcal{S}^{*}$.
\State By performing $K$-means clustering with the fingerprints of $\mathcal{P}^{*}$, the $l \times m$ temporally proposed particles are grouped into $K$ non-overlapping clusters $\mathcal{C}_1, \ldots, \mathcal{C}_K$.
\State Calculate the cluster-level surrogate likelihood as $W_{\mathrm{GBM}}(\mathcal{C}_k) = (1/|\mathcal{C}_k|) \sum_{i \in \mathcal{C}_k} w_{\mathrm{GBM}}(s_i^{*})$.
\State Resample $p/2$ particles from $\mathcal{P}^{*}$, denoted by $\mathcal{S}_0^*$, such that the number of particles in each $\mathcal{C}_k$ is proportional to $W_{\mathrm{GBM}}(\mathcal{C}_k)$.
\State Generate the remaining $p/2$ particles, $\mathcal{S}_1^*$, with equal probabilities, from the entries in $\mathcal{A}$.
\State Set the new particles as $\mathcal{P}_{t} = \mathcal{S}_0^* + \mathcal{S}_1^*$.
\State Calculate the likelihood $\mathcal{W}_t =\{w_{i}^t \}_{i=1}^p$ of all entries in $\mathcal{P}_t$.
\State The indices of the current particles are removed from the active candidate set $\mathcal{A}$.
\EndFor
\State Return $\mathcal{P}_t$ and $\mathcal{W}_t$ ($t=1, \ldots,T$) to calculate the approximate posterior Eq. \ref{eq:1}.
\end{algorithmic}
\end{algorithm}

\subsection{Ranking and prioritization}
In this study, we demonstrated that the SMC algorithm generally discovers an excessive number of hypothetical routes to a given product; in many cases, several hundreds or more for single-step reactions. The majority of such candidates are chemically unrealistic and false discoveries, which possibly results from the unavailability of failed reactions or data of low-yielding reaction in the machine-learning workflow. Indeed, our expert chemists recognized 60\% of the detected reactant pairs as having exceedingly low or no reactivity (Table S9). Here, we consider two means of prioritizing more promising candidates by using a heuristic ranking method or reaction-type grouping.

The ranking method scores a given pair of reactants $S$ as
\begin{eqnarray}
\gamma(S) = \alpha(Y_S) \max\{p(\{Y_S, S\} \in C_1), \ldots, p(\{Y_S, S\} \in C_{10})\}. \nonumber
\end{eqnarray}
The first term $\alpha(S)$ represents the probability of the tokenized SMILES symbols, which is provided by Molecular Transformer. In the second term, $p(\{Y_S, S\} \in C_i)$ indicates the probability of $\{Y_S, S\}$ belonging to a prescribed known reaction class $C_i$. In this study, we consider 10 reaction classes defined by Schneider and coworkers \cite{Schneider2016WhatsAssignment}, as described in Table S1. A total of 50k reactions in the United States Patent and Trademark Office (USPTO) dataset \cite{DanielMarkLowe2012ExtractionLiterature} were categorized into the 10 reaction classes. A sparse logistic regression model \cite{Friedman2009Glmnet:Models} was trained on a randomly selected 80\% portion of the dataset, in which the ECFPs of $Y_S$ and $S$ were calculated with diameter 4 and concatenated to obtain a descriptor. The prediction accuracy reached 97.3\% on the test set. The trained model was used to evaluate $p(\{Y_S, S\} \in C_i)$ and the most probable reaction class was selected to define the score. The underlying concept of using the hand-designed heuristic score was that candidate reactions that were highly discriminable into one of the known reaction classes were expected to be reliable.

Another method to choose the promising candidates is based on the grouping of reaction patterns. In this study, t-SNE \cite{VanDerMaaten2008VisualizingT-SNE} was performed on the augmented fingerprints of all given $S$, and X-means clustering \cite{DauPelleg2000X-means:2000} was used to automate the determination of the number of clusters and grouping of reaction patterns. For each cluster, we selected a representative reaction that exhibited the best within-cluster score. In this manner, we could infer how many different types of synthetic routes potentially existed or were feasible to design with a given set of purchasable compounds.

\section{Results}
\subsection{Data}

We used a collection of 50k single-step reactions \cite{Schneider2016WhatsAssignment} that were extracted from nine million patent applications and issued patents from 1976 to 2016. This dataset has served as a benchmark set in existing studies to evaluate retrosynthesis methods\cite{Liu2017RetrosyntheticModels, Coley2017Computer-AssistedSimilarity, Zheng2019PredictingNetworks, Lin2019AutomaticModels}. In this study, the dataset was used to train a forward model and to create the ground-truth sets of the single-step and two-step reactions, as described below. Following a previous study\cite{Liu2017RetrosyntheticModels}, we split the dataset into 80\% for training, 10\% for validation, and 10\% for testing.
Note that the recorded reactions were classified into 10 reaction classes according to an expert system \cite{Schneider2016WhatsAssignment} (Table S1). While most existing methods have employed one of the pre-defined reaction classes as input for the retrosynthetic prediction system to narrow it down to a limited solution space, we performed the Bayesian algorithm with no such prior knowledge. The solution space $\mathcal{S}$ that we considered hypothetically as a pool of purchasable compounds was spanned by all possible combinations of approximately 600k reactants in the USPTO dataset.

\subsection{Forward prediction model}
We used a pre-trained Molecular Transformer \cite{Schwaller2019MolecularPrediction}, which was trained on a dataset from USPTO. This attention-based neural translation model defined a translation between the SMILES strings of reactants and their product. For the sake of simplicity, the reagents were removed from the input. Any number of reactants, which were separated by ``.'', took part in the input. All of the reactions were canonicalized using RDKit. The inputs were tokenized with the regular expression according to a previous study \cite{Schwaller2019MolecularPrediction}. With the direct application to our test set, the distributed pre-trained model achieved top-1 accuracy of 70.7\% and top-5 accuracy of 86.4\%. However, with a fine-tuned model using the training and validation data, the top-1 accuracy reached 86.9\% and the top-5 accuracy reached 95.5\%. 

\subsection{Implementation}
The Bayesian retrosynthesis algorithm was implemented in Python (version 3.6.8), coupled with RDKit and sckit-learn. Molecular Transformer built using PyTorch was plugged into the forward model \cite{Schwaller2019MolecularPrediction}. All experiments were run on the AI Bridging Infrastructure (ABCI) at National Institute of Advanced Industrial Science and Technology. The computation was performed on executions with 25 nodes and 100 NVIDIA Tesla V100 devices.

\subsection{One-step retrosynthesis}
For the one-step retrosynthesis prediction, we randomly selected 100 test reactions from the test set consisting of one or two reactants (all of the reaction SMILES are presented in Table S7). Among the 100 reactions, Molecular Transformer could predict the true products to be top-1 candidates for 87 reactions. For each reaction tested, we performed the Bayesian retrosynthesis algorithm 10 times to evaluate the average detection rate. The number of particles was set to 1,000. In the first 100 steps of the SMC, each particle consisted of one reactant to explore the one-reactant reaction space. The reaction space of two reactants was explored in the subsequent 500 steps, in which a combination of two reactants constituted a particle. The SMC algorithm was used to perform a total of 600,000 ($=600 \times 1,000$) evaluations of the forward reaction models in each test reaction, which corresponded to approximately 0.0001\% of the entire search space. The Bayesian retrosynthesis algorithm identified multiple synthetic routes ending with 98.4\% of the 100 target molecules. Ground-truth reactants were found in 88.3\% of all test reactions. Focusing on the 87 reactions in which Molecular Transformer was predictable, the ground-truth reactants were found with a 94.5\% success rate (Table \ref{table:SMC_efficiency}).

\begin{table}[t]
\centering
\small
\begingroup
\renewcommand{\arraystretch}{1.3}
\textsf{
\begin{tabular}{lccc} \toprule
 & \begin{tabular}{c} Number of \\reactions \end{tabular} & \begin{tabular}{c}Detection of reactants ending  \\with target product [\%] \end{tabular} & \begin{tabular}{c}Inclusion of ground-truth \\reactants [\%]\end{tabular} \\ \hline
Random100             & 100                 & 98.4                   & 88.3                         \\
MT-predictable           & 87                  & 99.1                   & 94.5                   \\
\bottomrule
\end{tabular}
}
\endgroup
\caption{\footnotesize Performance of surrogate-accelerated SMC on randomly selected 100 ground-truth reactions (``Random100''). ``MT-predictable'' denotes the subset (87 reactions) in which Molecular Transformer could forwardly predict their products as top-1 candidates. The average success rate for detecting one or more synthetic routes ending with each target product is indicated in the third column. The fourth column denotes the average rate at which the ground-truth reactions were included in all detected routes.}
\label{table:SMC_efficiency}
\end{table}

\begin{figure}
  \centering
  \includegraphics[clip, width=16.5cm]{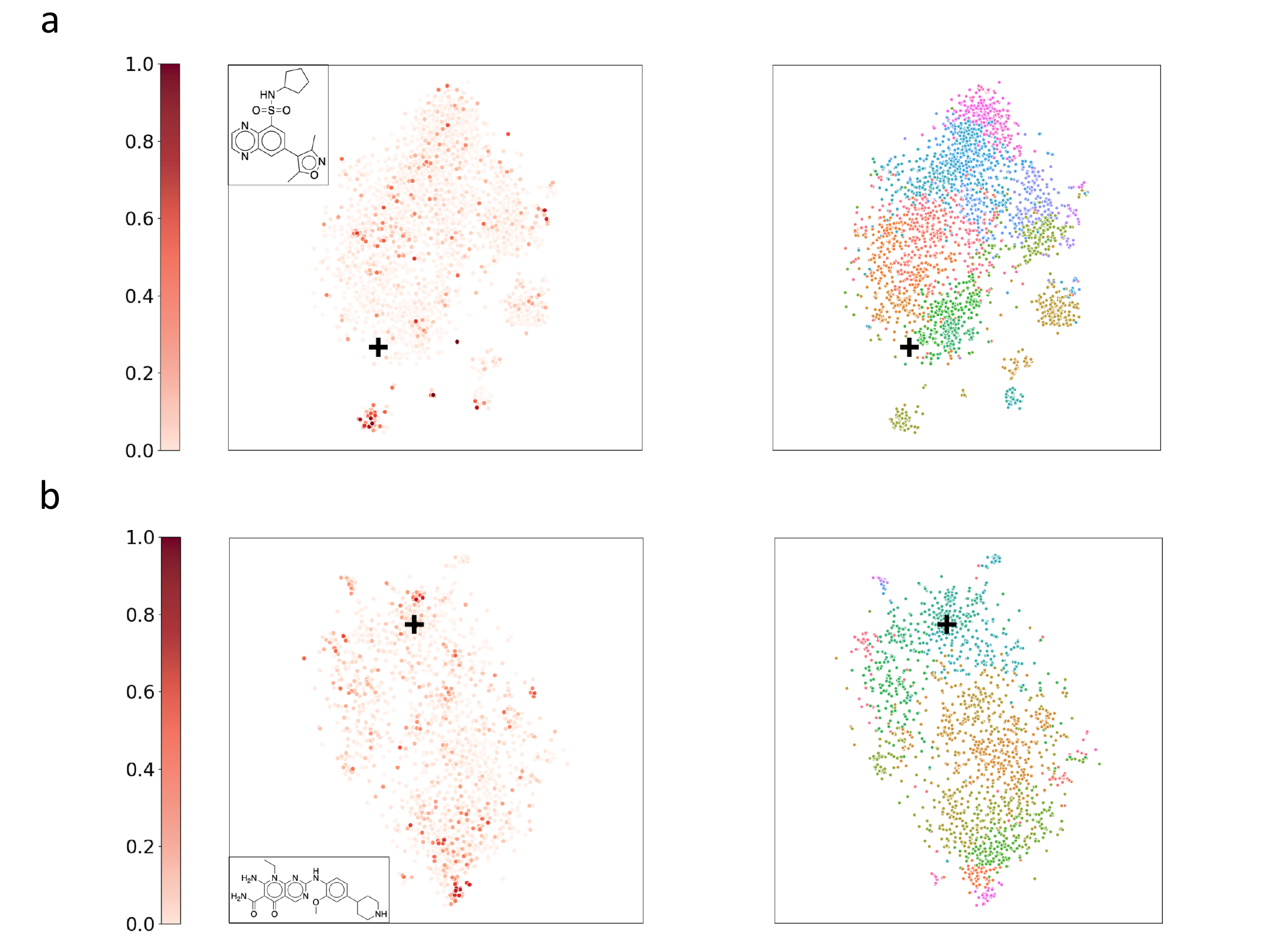}
  \caption{Distributions of candidate reactants for two synthetic targets (\textbf a and \textbf b) visualized based on projection to 2D space using t-SNE. {\textbf a.} The t-SNE projection of 3,559 candidate reactants ending with the target molecule (indicated in the top left corner) is shown. Here, $+$ denotes the ground truth reaction. In the left panel, the data points are color-coded according to the ranking scores normalized to $[0, 1]$. In the right panel, the X-means clustering of candidate reactants fingerprinted by ECFP. The optimal number of clusters is determined as 29. The cluster memberships are indicated with different colors in the t-SNE plot. {\textbf{b}} The t-SNE projection and X-means clustering of the other target molecule. A total of 2,236 candidate reactants are grouped into the 36 clusters.
  }
  \label{fig:singlestep}
\end{figure}

In this experiment, the Bayesian retrosynthesis algorithm revealed more than 400 synthetic routes on average for each design target.
Figure \ref{fig:singlestep} presents the diversity of the detected reaction routes on two synthetic targets. The ECFPs $\phi(S_1), \ldots, \phi(S_m)$ of $m$ reactants in $S$ were reduced to an augmented descriptor as $\phi(S) = \sum_{i=1}^m \phi(S_i)$, and these were embedded in the two-dimensional subspace for visualization using the t-SNE algorithm.
The distribution of the projected reactants indicates that there would be multiple motifs of the synthetic routes to the same product. The comprehensive detection of the candidate synthetic routes and their visualization-based summary may be helpful for facilitating chemists' creativity and decision-making in synthetic route design.

It is possible that most of such large amounts of candidate routes would be false findings. To narrow down the candidates, we performed the ranking procedure as described in the Methods section. For each of the 100 ground-truth reactions, we obtained a ranked list of the top-$N$ most probable candidates and investigated whether to include the ground-truth reactants. As summarized in Table \ref{table:accuracy_without_class}, the top-$N$ accuracy of the Bayesian retrosynthesis outperformed the state-of-the-art models for any $N \in \{1, 3, 5, 10\}$; in particular, 77.0\% and 80.3\% of the top-5 and top-10 accuracies were reached, respectively. Within the 87 ground-truth reactions in which Molecular Transformer successfully predicted the products in the forward manner, the top-5 and top-10 accuracies reached 86.0\% and 89.4\%, respectively.

According to the comparison presented in Table \ref{table:accuracy_without_class}, this result also outperformed other existing methods using the reaction class explicitly as an extra input in the retrosynthetic prediction. With a given reaction class, such prediction models can narrow down the reaction space into a focused region to enhance the prediction performance. However, such prior knowledge is rarely available in real applications. As a reference, we performed a modified Bayesian retrosynthesis. Given a true reaction class $C_i$ and its class probability $p(\{Y_S, S\} \in C_i)$, we calculated the ranking score as $\gamma(S) = \alpha(Y_s) p(\{Y_S, S\} \in C_i)$ instead of taking the maximum with respect to the classification probabilities, as shown in the Methods section. The top-5 and top-10 accuracies were further improved to 81.4\% and 83.5\%, respectively, for the 100 ground-truth reactions (Table \ref{table:accuracy_without_class}).

\begin{table}[t]
\centering
\small
\begingroup
\renewcommand{\arraystretch}{1.3}
\textsf{
\begin{tabular}{llcccc} \toprule
Data & Model & top-1 & top-3 & top-5 & top-10 \\ \hline
\multirow{5}{*}{Without reaction class} & similarity (Coley et al. 2017) \cite{Coley2017Computer-AssistedSimilarity} & 37.3                      & 54.7                      & 63.3                      & 74.1                      \\
& SCROP (Zheng et al. 2019) \cite{Zheng2019PredictingNetworks} & 43.7                      & 60.0                      & 65.2                      & 68.7                      \\
& Lin et al. 2020 \cite{Lin2019AutomaticModels} & 43.1                      & 64.6                      & 71.8                      & 78.7                      \\
& Bayesian-Retro                  & \textbf{47.5}             & \textbf{67.2}             & \textbf{77.0}             & \textbf{80.3}             \\ \cdashline{2-6}
& Bayesian-Retro (MT-predictable) & 54.6                      & 74.9                      & 86.0                      & 89.4                      \\ \hline
\multirow{7}{*}{With reaction class} & baseline (Liu et al. 2017) \cite{Liu2017RetrosyntheticModels} & 35.4                      & 52.3                      & 59.1                      & 65.1                       \\
& seq2seq (Liu et al. 2017) \cite{Liu2017RetrosyntheticModels} & 37.4                      & 52.4                      & 57.0                      & 61.7                       \\
& similarity (Coley et al. 2017) \cite{Coley2017Computer-AssistedSimilarity} & 52.9                      & 73.8                      & 81.2                      & \textbf{88.1}                       \\
& SCROP (Zheng et al. 2019) \cite{Zheng2019PredictingNetworks} & \textbf{59.0}                      & \textbf{74.8}                      & 78.1                      & 81.1                       \\
& Lin et al. 2020 \cite{Lin2019AutomaticModels} & 54.6                      & \textbf{74.8}                      & 80.2                      & 84.9                       \\
& Bayesian-Retro    & 55.2                      & 74.1                      & \textbf{81.4}                      & 83.5                       \\ \cdashline{2-6}
& Bayesian-Retro (MT-predictable) & 62.8             & 81.8             & 89.7             & 91.7              \\
\bottomrule
\end{tabular}
}
\endgroup
\caption{\footnotesize Performance of various retrosynthetic prediction methods with or without reaction class labels. ``Bayesian-Retro'' denotes the proposed method; ``MT-predictable'' denotes the performance on the 87 ground-truth reactions with their products forwardly predictable by Molecular Transformer. The top-1, top-3, top-5, and top-10 accuracies in [\%] are indicated for each case.}
\label{table:accuracy_without_class}
\end{table}

\subsection{Multi-step retrosynthesis}

The Bayesian retrosynthesis algorithm was applied to the design of two-step synthetic routes. By combining the single-step reactions that were predictable with Molecular Transformer in the test reaction set,
we generated a ground-truth set of two-step reactions as follows: If a product of a recorded reaction appeared in a different reaction as a reactant, the two reactions were connected to form a two-step synthetic route, as the product of the former reaction was involved as a reactant of the second-step reaction as an intermediate product. In this manner, we obtained 11 two-step reactions. Our expert chemists verified the validity of these reactions, as summarized in Table \ref{table:multistep} and Table S8. 
According to their evaluations, in which unrecorded reagents and reaction conditions were inferred based on expert knowledge, the first step in reaction 3 was judged as chemically unrealistic. In this case, instead of excluding reaction 3 from the ground-truth set, we tested whether the Bayesian retrosynthesis could determine alternative synthetic routes to the target product.

The Bayesian retrosynthesis algorithm was performed 10 times for each reaction. The number of particles was set to 2,000, and each particle consisted of two and one reactants in the first and second reactions, respectively. The number of steps in the SMC was set to 1,000, constituting a total of $2 \times 10^6$ searches in each test case. This number corresponded to approximately $10^{-11}$ of the entire search space. In 9 of the 11 reactions, the recorded reactants could successfully be identified at least once among the 10 repeated tests. More than 11,000 candidate routes were identified on average for each target product. 
Hence, we aggregated all of the candidate synthetic routes and performed the ranking procedure. 
The recorded reactions were ranked as the top 10 candidates in five cases and as the top 100 in the other cases (Table \ref{table:multistep}).

\begin{figure}
  \centering
  \includegraphics[clip, width=16.5cm]{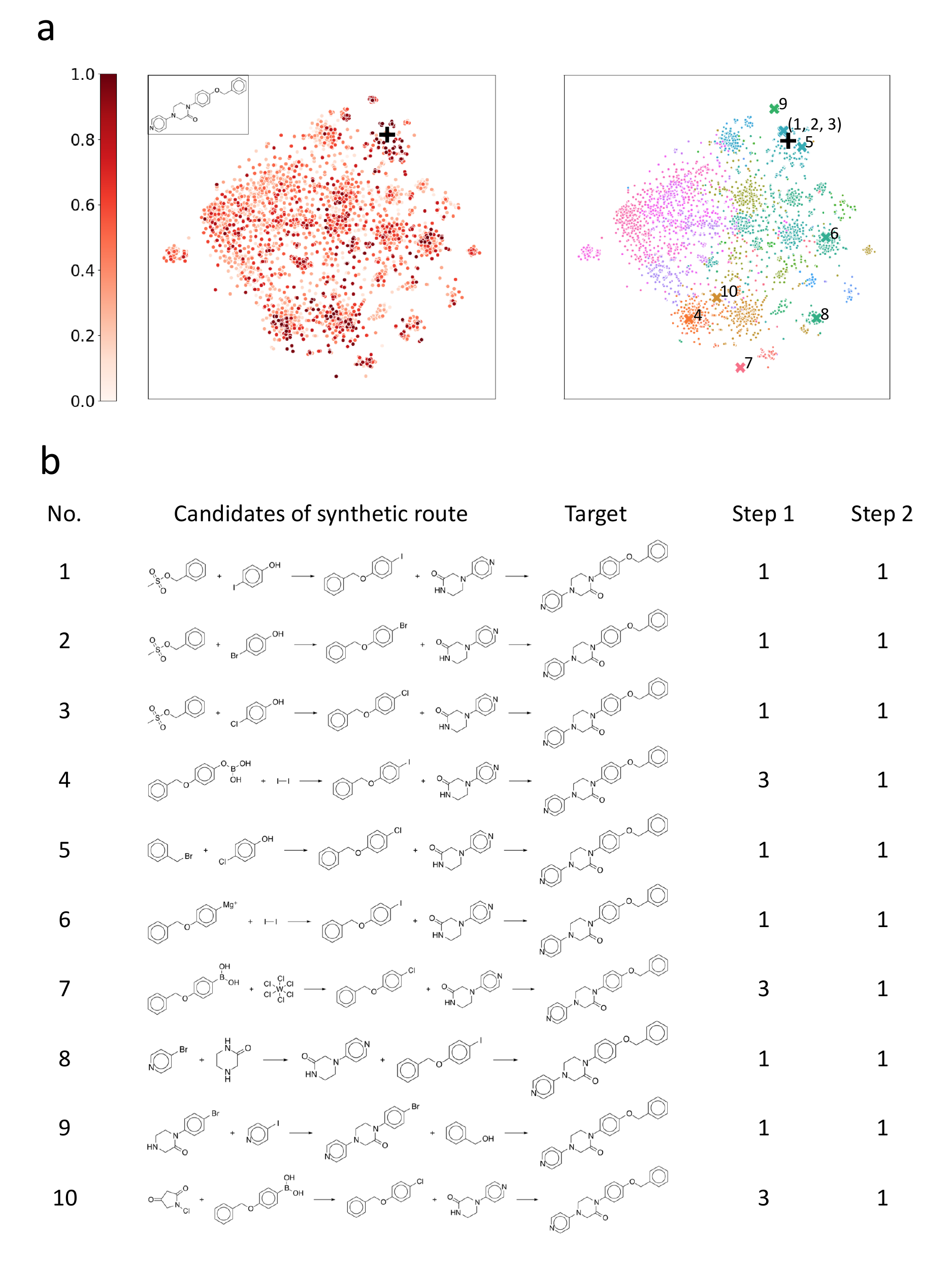}
  \caption{{\textbf a.} t-SNE projection of 6,613 candidates for two-step synthetic routes to target product in reaction 9, where $+$ denotes the ground-truth reaction route. In the left panel, the data points are color-coded according to the scores normalized to $[0, 1]$. In the right panel, the X-means clustering classified the 6,613 candidate routes into 98 groups, which are mapped on the t-SNE plot. The identified clusters are indicated in different colors. $\times$ denotes the 10 candidate routes presented in \textbf{b}.
  {\textbf b.} 10 candidate routes belonging to different clusters. A score $\{1, 2, 3\}$ is assigned to each reaction step, indicating `feasible,' `contentious' or `infeasible,' as judged by expert chemists.}
  \label{fig:multistep}
\end{figure}

To observe the distribution of the candidate synthetic routes, the t-SNE projection of the detected reactants for reaction 9 is presented in Figure \ref{fig:multistep}a. Candidate reactants closer to the recorded reactants exhibited higher scores. To identify the different motifs in the candidate synthetic routes, X-means clustering was applied to the ECFPs of the reactants, which were grouped into 98 clusters. We investigated the synthesis feasibility of the 10 reactants exhibiting the highest score in each of the 10 different clusters (Figure \ref{fig:multistep}a). According to the evaluations by the expert chemists, 7 out of the 10 proposed routes would be chemically reactive and synthesizable. In candidate routes 1 to 3, the first and second steps were known as Williamson ether synthesis \cite{QJ8520400229} and a palladium-catalyzed coupling reaction \cite{Miyaura1995Palladium-CatalyzedCompounds}, respectively. In candidate route 9, the second step consisted of ether synthesis. It should be stressed that ranking only by the score does not always reveal such promising synthetic routes. It is important to extract a diverse set of candidates based on the clustering procedure to enhance chemists' creativity.

\begin{figure}
  \centering
  \includegraphics[clip, width=16.5cm]{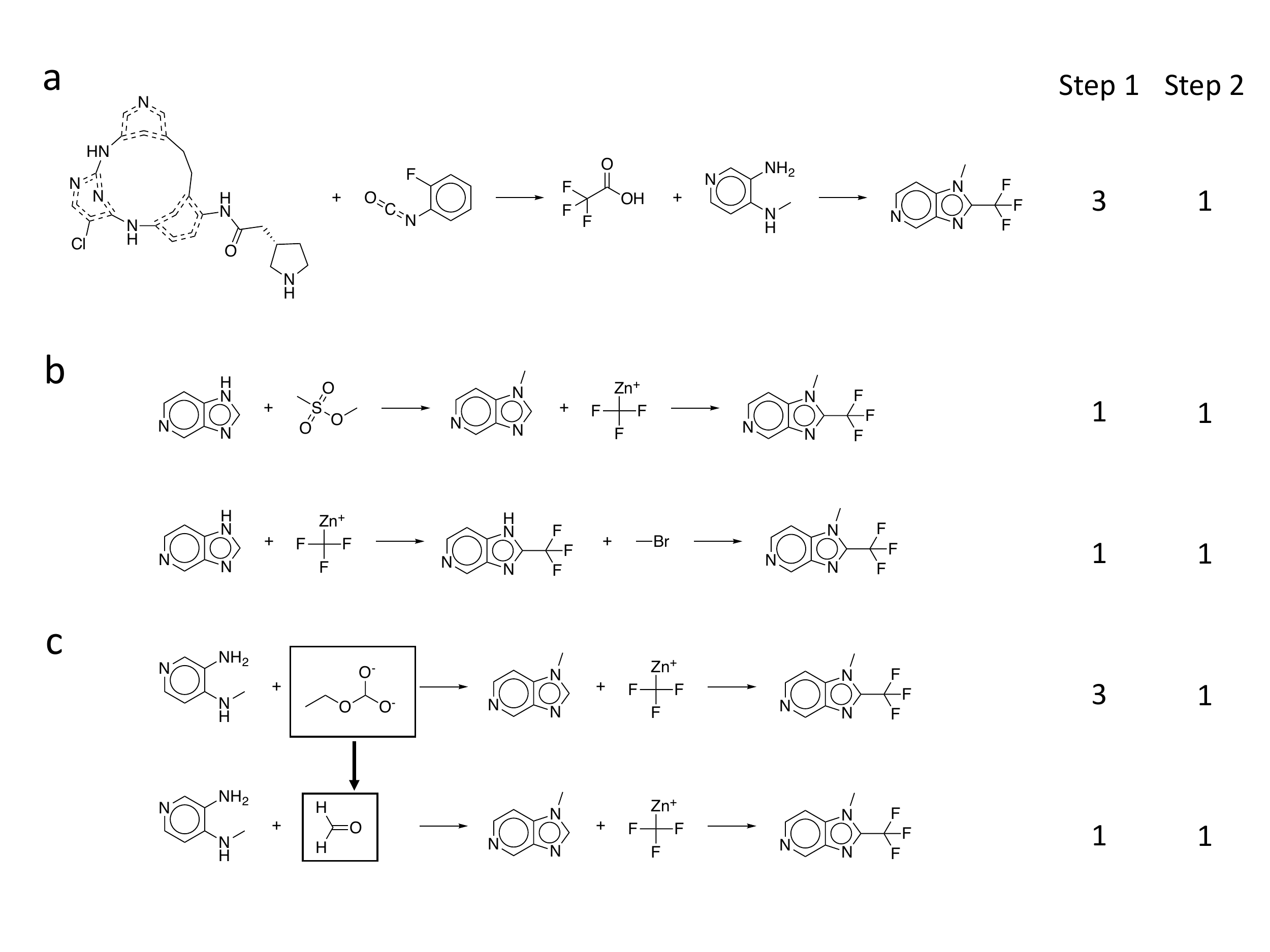}
  \caption{{\textbf a.} Two-step reaction 3 in test dataset. The score $\{1, 2, 3\}$ denotes chemists' judgments on the synthesis feasibility.  {\textbf b.} Two alternative synthetic routes proposed by Bayesian retrosynthesis algorithm. {\textbf c.} The top reaction is a ring-forming synthetic route proposed by Bayesian retrosynthesis algorithm. The first step is infeasible because the orthoformic acid monoester is chemically unstable. The bottom reaction is a ring-forming synthetic route suggested by chemists based on the proposed synthetic route indicated at top.}
  \label{fig:reaction3}
\end{figure}

For ground-truth reaction 3, which the chemists judged to be chemically infeasible, the Bayesian retrosynthesis algorithm identified 2,280 alternative synthetic routes to the target molecule (Figure \ref{fig:reaction3}). Based on our ranking and clustering procedures, 10 synthetic routes were selected and two reactions with different side-chain modifications were judged as reactive and synthesizable according to chemists' evaluations (Figure \ref{fig:reaction3}b). Moreover, a ring-forming synthesis was proposed by the algorithm (the route indicated at the top of Figure \ref{fig:reaction3}c). Although the orthoformic acid monoester used in the first step was chemically unstable, the proposed synthetic route could be helpful for chemists to consider the strategy of designing alternative synthetic routes. Indeed, a different ring-forming synthetic route was manually designed by using formaldehyde instead of the orthoformic acid monoester.

\begin{landscape}

\begin{table}[ht]
\caption{\footnotesize Ground-truth set consisting of two-step synthetic routes to 11 targets. The score $\{1, 2, 3\}$, which is assigned to each reaction step, indicates `feasible,' `contentious' or `infeasible,' as judged by the chemists. The Bayesian retrosynthesis was performed 10 times for each reaction. The fifth column denotes the total number of candidate routes ending with the target molecule in the 10 trials. T and F denote the presence or absence of the recorded reaction in the detected synthetic routes. The final column indicates the rank of the ground truth predicted by the ranking algorithm.}
\label{table:multistep}
\centering
\small
\begingroup
\renewcommand{\arraystretch}{1.3}
\textsf{
\begin{tabular}{lcccccc} \toprule
No. & Reaction & Step 1 & Step 2 &
\begin{tabular}{c}
No. of\\
candidates
\end{tabular} &
\begin{tabular}{c}
Ground \\
truth
\end{tabular} &
Rank\\ \hline
1 &
\begin{minipage}{0.56\hsize}
\centering
\includegraphics[clip,width=0.90\hsize]{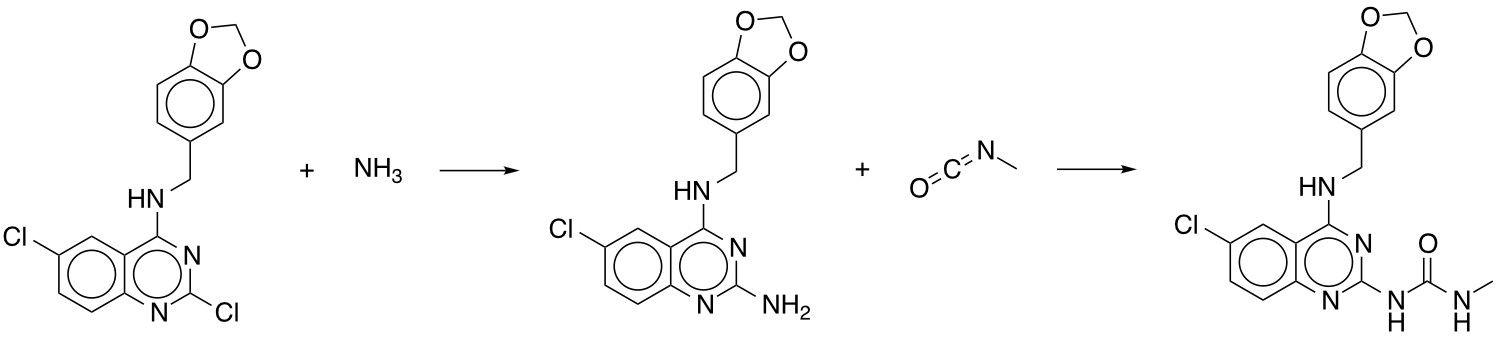}
\end{minipage} &
1 & 1 &
784 & F & -\\
2 &
\begin{minipage}{0.56\hsize}
\centering
\includegraphics[clip,width=0.90\hsize]{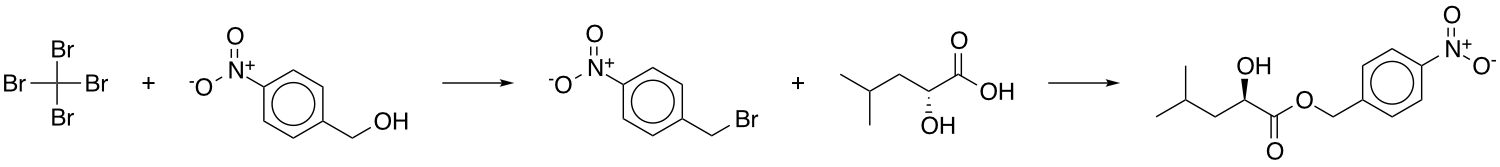}
\end{minipage} &
1 & 1 &
1,326 & T & 2\\
3 &
\begin{minipage}{0.56\hsize}
\centering
\includegraphics[clip,width=0.90\hsize]{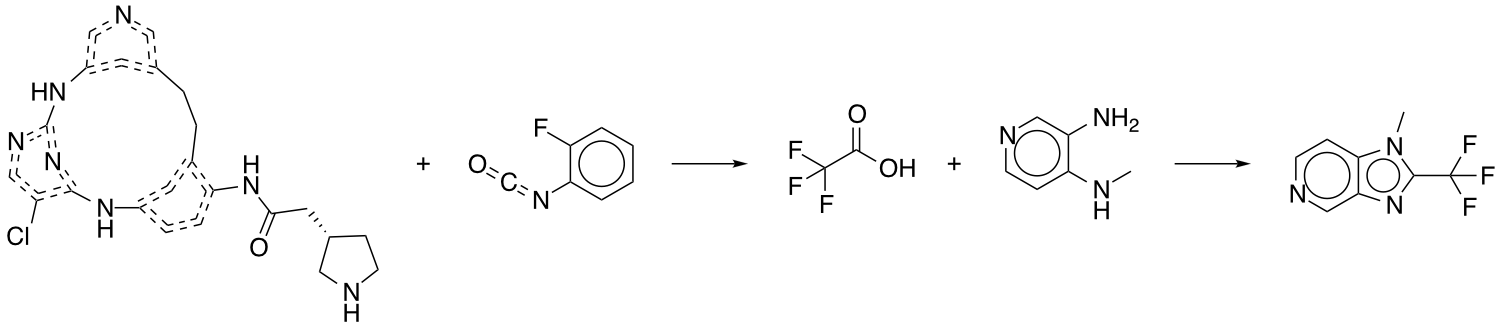}
\end{minipage} &
3 & 1 &
2,280 & F & -\\
4 &
\begin{minipage}{0.56\hsize}
\centering
\includegraphics[clip,width=0.90\hsize]{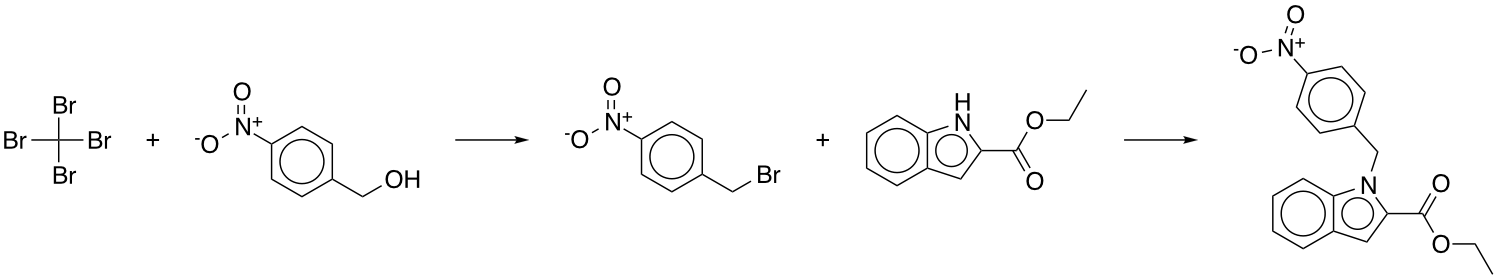}
\end{minipage} &
1 & 1 &
18,197 & T & 5\\
5 &
\begin{minipage}{0.56\hsize}
\centering
\includegraphics[clip,width=0.90\hsize]{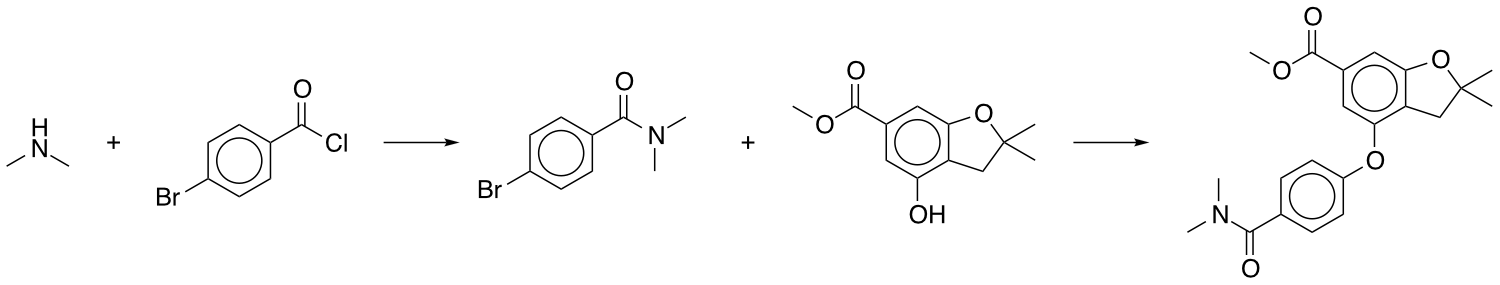}
\end{minipage} &
1 & 1 &
28,388 & T & 94\\
\bottomrule
\end{tabular}
}
\endgroup
\end{table}

\begin{table}[ht]
\centering
\small
\begingroup
\renewcommand{\arraystretch}{1.3}
\textsf{
\begin{tabular}{lcccccc} \toprule
No. & Reaction & Step 1 & Step 2 &
\begin{tabular}{c}
No. of\\
candidates
\end{tabular} &
\begin{tabular}{c}
Ground \\
Truth
\end{tabular} &
Rank\\ \hline
6 &
\begin{minipage}{0.56\hsize}
\centering
\includegraphics[clip,width=0.90\hsize]{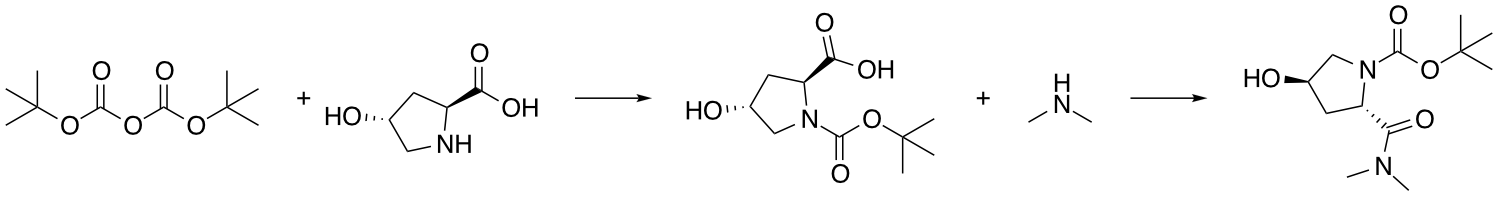}
\end{minipage} &
1 & 1 &
48,601 & T & 3\\
7 &
\begin{minipage}{0.56\hsize}
\centering
\includegraphics[clip,width=0.90\hsize]{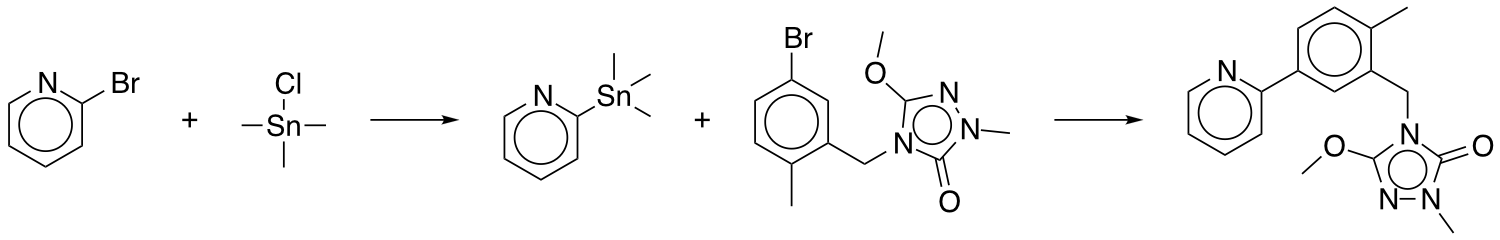}
\end{minipage} &
1 & 1 &
7,622 & T & 45\\
8 &
\begin{minipage}{0.56\hsize}
\centering
\includegraphics[clip,width=0.90\hsize]{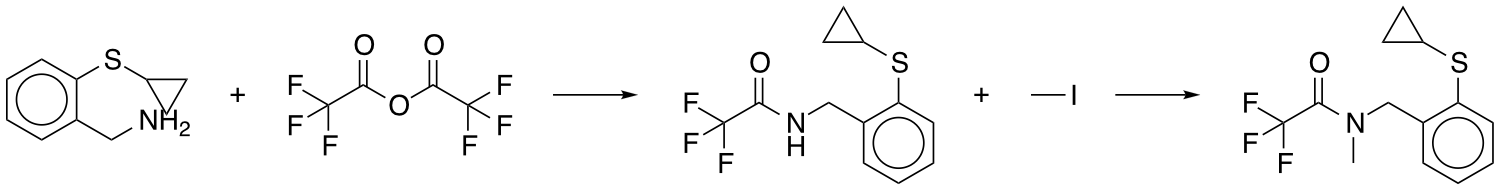}
\end{minipage} &
1 & 1 &
2,167 & T & 17\\
9 &
\begin{minipage}{0.56\hsize}
\centering
\includegraphics[clip,width=0.90\hsize]{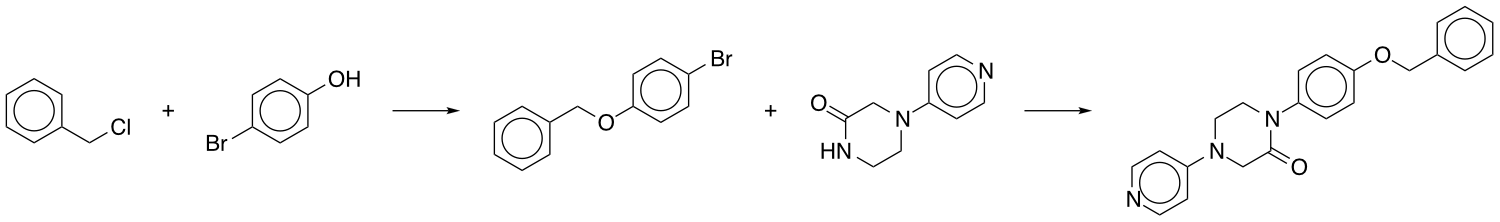}
\end{minipage} &
1 & 1 &
6,613 & T & 6\\
10 &
\begin{minipage}{0.56\hsize}
\centering
\includegraphics[clip,width=0.90\hsize]{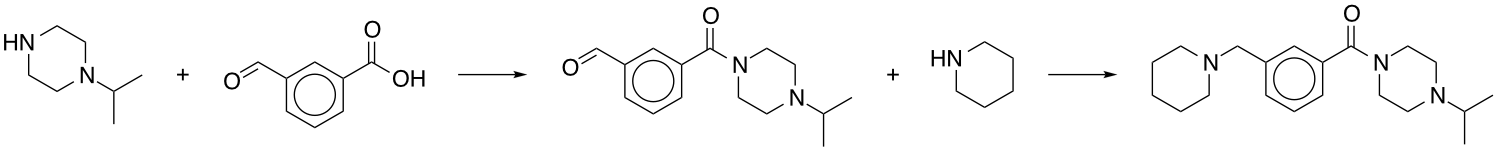}
\end{minipage} &
1 & 1 &
3,122 & T & 23\\
11 &
\begin{minipage}{0.56\hsize}
\centering
\includegraphics[clip,width=0.90\hsize]{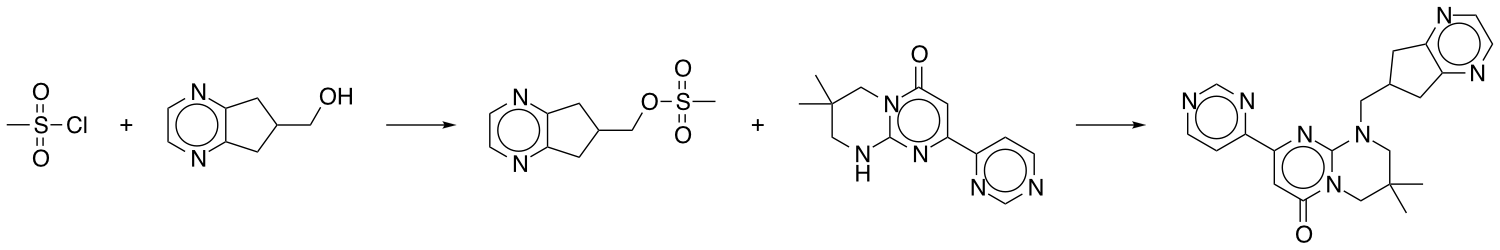}
\end{minipage} &
1 & 1 &
4,838 & T & 8\\
\bottomrule
\end{tabular}
}
\endgroup
\label{table:multistep_continue}
\end{table}

\end{landscape}

\section{Conclusions}

We developed a Bayesian retrosynthesis algorithm. Most previous studies have focused on the direct prediction of the reaction inputs backwardly from a target product. In general, the backward prediction task is significantly more difficult than that of forward prediction, as the model needs to reconstruct several building blocks of reactant molecules that are generally missing in the target product. Moreover, most reactants resulting from such backward prediction will not be available for purchase and the candidate itself becomes a synthetic target. To overcome these obstacles, we firstly overhauled the problem of retrosynthetic prediction. We obtained a forward reaction model to define the mapping from reactants to products, achieving a high level of accuracy. Thereafter, the retrosynthetic prediction was addressed by exploring the inverse mapping from a target product to a pair of reactants in the given forward model, in which all possible pairs of purchasable compounds spanned the feasible solution space. The prediction accuracy on benchmark datasets outperformed the current state-of-the-art methods, as 80.3\% and 50\% of the known test reactions were successfully rediscovered within top-10 accuracy for the single-step and two-step synthetic route planning, respectively.


The Bayesian retrosynthesis algorithm revealed the presence of numerous alternative routes towards the target product, which were programmed in the trained reaction prediction model. The identification of such diverse candidate routes may be helpful for chemists to facilitate their creative works in synthetic organic chemistry. However, our expert chemists concluded that nearly 60\% of the proposed two-step reactions would be false discoveries owing to no or exceedingly low reactivity. The prediction of the presence or absence of reactivity is currently beyond the capabilities of any synthetic prediction models, because these are trained only on instances from highly reactive reactions in the published data. The lack of negative data on failed reactions or low-yielding reactions prevents us from obtaining machine-learning models to discriminate the presence or absence of reactivity in a candidate synthetic route. Several previous studies generated artificial negative examples by perturbing and shuffling the reported known reactions. In this study, we introduced a heuristic rule for the ranking and prioritization of candidate reaction routes. However, none of these methods can solve the problem at a fundamental level, and eventually, we have to create a comprehensive dataset of negative reactions from experimental observations in laboratory synthesis, the literature, chemists' hand-coded heuristics or high-throughput quantum chemistry calculations.

\section{Acknowledgement}
This work was supported in part by the Materials Research by Information Integration Initiative (MI$^2$I) of the Support Program for Starting Up Innovation Hub from the Japan Science and Technology Agency (JST). Ryo Yoshida acknowledges the financial support received from a Grant-in-Aid for Scientific Research (A) 19H01132 from the Japan Society for the Promotion of Science (JSPS), JST CREST Grant Number JPMJCR19I1, Japan, and JSPS KAKENHI Grant Number 19H05820. Stephen Wu acknowledges the financial support received from JSPS KAKENHI Grant Number JP18K18017.

\newpage
\begin{suppinfo}

\begin{itemize}
  \item \textbf{Figure S1}: Detected candidates on two-step reaction routes to target products in reactions 1, \ldots, 11.
  \item \textbf{Table S1}: Reaction classes used to label 50k reactions and numbers of reactions in each class.
  \item \textbf{Table S2}: Top-10 accuracy of various retrosynthetic prediction methods when reaction class labels are given.
  \item \textbf{Tables S3, S4 and S5}: Parameters and experimental conditions for simple SMC, surrogate-assisted SMC (one-step) and surrogate-assisted SMC (multi-step).
  \item \textbf{Table S6}: SMILES and detailed results of 40 reactions for performed test of simple SMC.
  \item \textbf{Table S7}: SMILES and detailed results of 100 reactions used in one-step retrosynthesis.
  \item \textbf{Table S8}: SMILES of 11 reactions used in multi-step retrosynthesis.
  \item \textbf{Table S9}: Detailed results of 11 reactions used in multi-step retrosynthesis.
\end{itemize}

\end{suppinfo}

\bibliography{references}

\end{document}